\documentclass[journal]{IEEEtran}
\usepackage{amsmath}
\usepackage{amssymb}
\usepackage{graphicx}
\usepackage{booktabs}
\usepackage{multirow}
\usepackage{enumitem}
\usepackage{capt-of}
\usepackage{tcolorbox}
\usepackage{xcolor}
\usepackage[framemethod=TikZ]{mdframed}
\usepackage{algorithm}
\usepackage{algorithmic}

\ifCLASSINFOpdf

\else

\fi

\begin{document}
%

\title{CovNav: Conformal Prediction-Guided Adaptive Inference for Object Goal Navigation}

\author{Pengying Wu, Zhejia Wen, Kang Ding, Kangjie Zhou, Chunxuan Jiao, Wei Pan,~\IEEEmembership{Member,~IEEE},\\ Xiangli Nie,~\IEEEmembership{Member,~IEEE}, and Chang Liu,~\IEEEmembership{Member,~IEEE}%

\thanks{P. Wu, Z. Wen, K. Ding, K. Zhou, C. Jiao, and C. Liu are with the School of Advanced Manufacturing and Robotics, Peking University, Beijing 100871, China (Corresponding author: Chang Liu.) (e-mail: changliucoe@pku.edu.cn). X. Nie is with the State Key Laboratory of Multimodal Artificial Intelligence Systems, Institute of Automation, Chinese Academy of Sciences, Beijing 100190, China (e-mail: xiangli.nie@ia.ac.cn). W. Pan is with the School of Engineering, Newcastle University, UK (e-mail: wei.pan2@newcastle.ac.uk).}}

\maketitle
\begin{abstract}
In object goal navigation (ObjectNav), an embodied agent must find a target object (e.g., ``find a bed'') in a previously unseen environment, relying on spatial commonsense such as object co-occurrence. Large Language Models (LLMs) encode this commonsense, but deploying them on robots exposes a trade-off between reasoning quality and efficiency: cloud-based models incur latency and privacy risks, whereas small on-device LLMs are fast but prone to hallucinated spatial judgments. Hybrid reasoning LLMs offer both a fast non-thinking mode and a slow thinking mode, which raises a key question for embodied agents: when is slow thinking worth its cost? We propose \textbf{CovNav}, a training-free framework that answers this question with Conformal Prediction (CP), turning dual-process (System 1 / System 2) inference into an uncertainty-aware decision. Much like recent decision models such as Jev, CovNav casts each spatial query as a structured binary decision rather than free-form generation: a local LLM in fast mode judges whether the target is likely near an observed object, CP converts its token probabilities into a prediction set, and only ambiguous queries are escalated to slow chain-of-thought reasoning. The object-pair formulation keeps calibration tractable and supports open-set objects through off-the-shelf detectors. Experiments on Gibson and HM3D show that CovNav surpasses the strongest zero-shot baseline by 0.8 and 1.7 percentage points in success rate on HM3D v1 and v2, while invoking slow reasoning for only 12.4\% of queries and retaining most of the gain of always-on thinking at a fraction of its latency. Real-robot experiments on a TurtleBot 4 further validate the approach. CovNav shows how calibrated uncertainty can decide when an embodied agent should think slowly, offering a principled form of adaptive test-time reasoning for LLM-based robots.

\vspace{1em}

\end{abstract}


\vspace{1em}

\IEEEpeerreviewmaketitle

\section{Introduction}

\begin{figure}[t]
\centering
\includegraphics[width=\columnwidth]{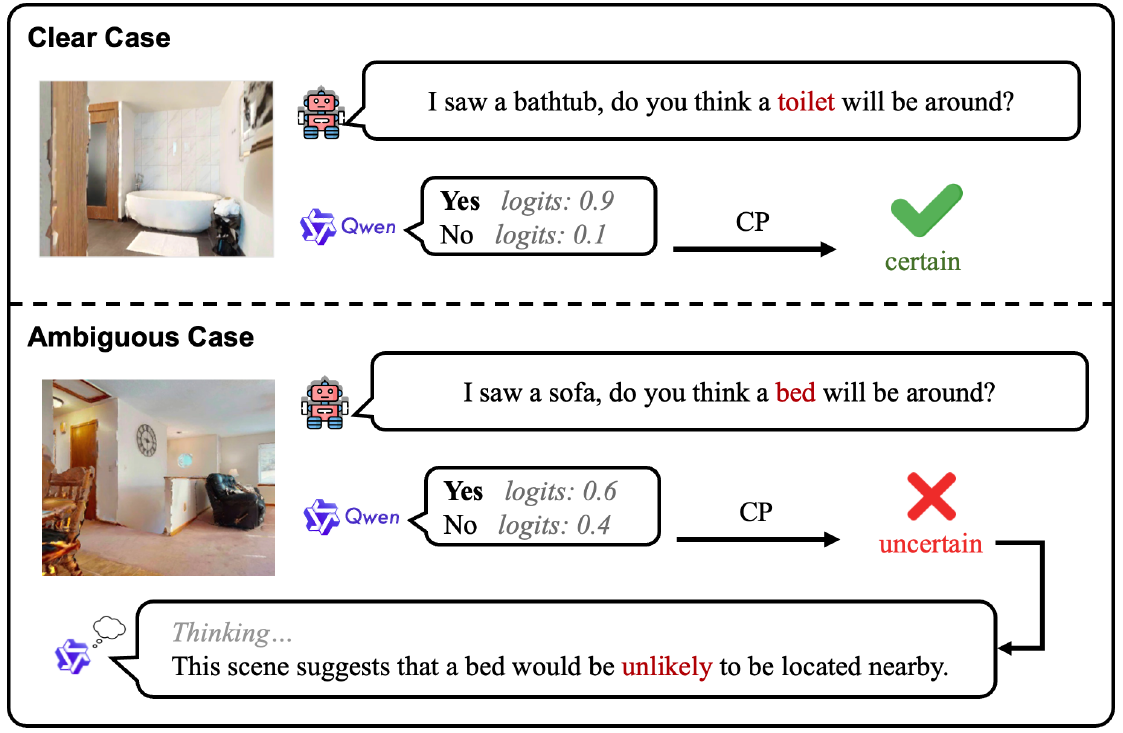}
\caption{We propose a Conformal Prediction (CP)-guided fast-slow inference paradigm for object goal navigation using Qwen$3$-$8$B as the local LLM. CP monitors the local LLM's output confidence to adaptively switch Qwen between its fast direct inference mode and slow deliberative reasoning mode. \textbf{Clear Case:} When observing a bathtub, Qwen confidently predicts a toilet will be nearby (logits: 0.9 vs 0.1), so CP accepts the fast direct inference. \textbf{Ambiguous Case:} When observing a sofa and querying about a bed, Qwen shows low confidence (logits: 0.6 vs 0.4), so CP triggers its slow deliberative reasoning mode for more reliable spatial analysis.}
\label{fig:teaser}
\end{figure}

\IEEEPARstart{R}{ecent} advances in large foundation models have achieved remarkable success across language understanding, visual reasoning, and knowledge-intensive tasks. A pivotal next step for artificial intelligence is to ground these capabilities in the physical world, building embodied agents that can perceive, reason, and act within complex, unstructured environments~\cite{11098567}. Among the core competencies required for such agents, the ability to autonomously navigate to semantically specified targets is both fundamental and practically significant, underpinning applications from domestic assistance to search and rescue.

Object goal navigation (ObjectNav) serves as a representative testbed for this capability. Given a semantic target (e.g., ``find a bed''), an agent must explore and navigate through a previously unseen environment without any prior map~\cite{chaplot2020object,fukushima2022object}. A key enabler for efficient exploration is commonsense spatial knowledge, that is, the understanding that functionally related objects tend to co-locate within the same room (e.g., beds and nightstands in bedrooms, sofas and televisions in living rooms). Large Language Models (LLMs), which encode vast commonsense knowledge from pre-training, have recently emerged as a powerful tool for providing such spatial priors without task-specific training~\cite{zhou2023esc,wu2024voronav,wen2025zero}, motivating a growing body of work on LLM-based object goal navigation.

Existing LLM-based navigation approaches face significant deployment challenges. First, many state-of-the-art methods rely on cloud-based APIs (e.g., ChatGPT, Claude)~\cite{wu2024voronav,cai2025cl}, which introduces network latency, raises privacy concerns, and limits deployment in offline or network-constrained scenarios. Second, directly deploying smaller open-source LLMs on resource-constrained robot platforms often leads to performance degradation, manifesting as hallucination and unreliable spatial judgments when processing complex relationships. Third, existing approaches lack adaptive mechanisms to dynamically adjust inference depth based on problem difficulty, resulting in either consistently invoking large models (incurring high computational costs) or relying exclusively on small models (sacrificing accuracy). This raises a critical question: \textit{How can we achieve both efficient and accurate spatial reasoning on resource-constrained robotic platforms?}

To address this challenge, we propose an adaptive inference framework that dynamically adjusts inference depth based on problem difficulty, utilizing LLM prediction uncertainty as a core metric. A key insight driving our work is that not all spatial reasoning problems require deep deliberation, as illustrated in Figure~\ref{fig:teaser}. For instance, judging the spatial proximity between a television and a sofa is straightforward, given their typical co-occurrence in living rooms. However, determining the relationship between a sofa and a bed requires cross-room reasoning to recognize they belong to distinct functional spaces. Crucially, the output confidence produced by small LLMs, specifically the logits generated during deployment, serves as an effective indicator of problem difficulty, reflecting how well the input aligns with the training corpus. We leverage Conformal Prediction (CP)~\cite{angelopoulos2021gentle}, a statistically rigorous uncertainty quantification framework, to assess this confidence without requiring additional model training. CP constructs prediction sets based on model outputs: when the model is highly confident, the prediction set contains a single answer, indicating that the fast direct inference mode is sufficient. Conversely, when uncertain, the prediction set contains multiple possible answers, indicating the model's inability to identify a unique solution, which triggers a slow deliberative reasoning mode. This mechanism mirrors human cognitive processing: System 1 thinking (fast, intuitive) handles routine decisions, while System 2 thinking (slow, analytical) engages for complex problems~\cite{kahneman2011thinking}.

This perspective also connects CovNav to two recent trends in LLM reasoning. First, hybrid reasoning models such as Qwen3~\cite{yang2025qwen3} expose both a fast non-thinking mode and a slow thinking mode, and recent work trains such models with reinforcement learning to decide \textit{when to think}~\cite{zhang2025adaptthink,tu2025learning}; CovNav makes this decision without any training, using a statistically calibrated criterion. Second, decision models such as Jev~\cite{almeida2026jev,ling2026jev} answer queries with structured choices, binary judgments, or scores instead of free-form text, trading open-ended generation for speed and reliability. CovNav's fast mode follows the same principle by casting each spatial query as a binary object-pair decision with a confidence score, and adds a CP gate that escalates to System 2 reasoning only when this System 1 decision is uncertain.

Building on this insight, we design a CP-guided navigation framework (CovNav) where the robot constructs an object-wise semantic map using off-the-shelf detection models. For each newly observed object, it queries the LLM to assess spatial relationships with the target object. CP analyzes the response confidence to determine whether the fast direct inference mode can reliably answer the question. If confidence is sufficient, the system directly utilizes the response; otherwise, it activates the slow deliberative reasoning mode for more thorough analysis. The resulting semantic value map, weighted by these assessments, guides the robot toward high-potential exploration regions. Two key design choices facilitate this framework: (1) formulating queries as object-pair relationships rather than open-ended visual question answering makes the problem space more tractable for CP calibration, and (2) leveraging open-source detection models enables open-set semantic understanding without task-specific training.

The main contributions of this work are as follows:

\begin{itemize}
\item We propose a novel fast-slow inference paradigm for robotic navigation that adaptively switches between fast direct inference and slow deliberative reasoning based on LLM response uncertainty, effectively achieving a balance between computational efficiency and decision accuracy on resource-constrained platforms.
\item We introduce the first use of Conformal Prediction to gate between the fast and slow reasoning modes of an LLM for object navigation, providing a theoretically grounded and training-free switching mechanism. We further design an object-pair-based query formulation that makes CP calibration tractable while enabling open-set semantic understanding through off-the-shelf vision models.
\item We validate our approach through extensive simulated and real-world robot experiments, demonstrating significant improvements in both navigation efficiency and success rate compared to existing methods.
\end{itemize}

The remainder of this paper is organized as follows. Section~\ref{sec:related} reviews related work in zero-shot object navigation, LLM-based navigation and uncertainty quantification in Embodied AI. Section~\ref{sec:method} provides the problem formulation and details our proposed methodology, specifically the object-based semantic mapping and the CP-guided fast-slow switching mechanism. Section~\ref{sec:experiments} presents extensive experimental results, including simulation comparisons and ablation studies. Section~\ref{sec:realworld} details real-world validations, demonstrating the effectiveness and efficiency of our approach in physical environments. Section~\ref{sec:conclusion} concludes the paper and discusses future directions.

\section{Related Work}
\label{sec:related}

\subsection{Zero-Shot Object Goal Navigation}
Object goal navigation (ObjectNav) requires an agent to navigate to instances of specific object categories in unknown environments, combining visual perception, semantic understanding, and spatial reasoning \cite{10475904, chen2023not, ginting2024seek}. It is closely related to other embodied navigation tasks, such as image-goal navigation \cite{lei2025gaussnav} and multi-agent cooperative navigation \cite{10636265}. End-to-end approaches learn navigation policies directly from observations via reinforcement learning \cite{10452415}, but tend to generalize poorly to unseen environments compared with modular pipelines \cite{chaplot2020object}. The Object Memory Transformer (OMT) \cite{fukushima2022object} introduced long-term semantic memory to improve navigation efficiency, while Goal-Oriented Semantic Exploration \cite{chaplot2020object} pioneered episodic semantic maps with efficient exploration strategies.
Recent modular approaches decompose navigation into separate components. The SEEK framework \cite{ginting2024seek} combines spatial configurations with commonsense knowledge about object relationships. Liu et al. \cite{11199871} explore implicit scene memory for target-directed navigation by storing historical observations in a NeRF-based memory and retrieving target-relevant cues through dual-space attention with uncertainty masking. Heterogeneous Scene Representation (HSR) \cite{10556584} models human-robot-object interactions using graph attention networks, while Commonsense-Aware Object Value Graph (OVG-Nav) \cite{10478188} incorporates topological structures leveraging visual observations and commonsense priors. Multi-View Visual Transformer (MVVT) \cite{10839297} expands input perspectives to capture richer spatial context relationships for navigation.

Zero-shot object navigation (ZSON) enables agents to locate objects without explicit training on specific categories or environments \cite{kuang2024openfmnav}, \cite{10629078}. ZSON \cite{majumdar2022zson} established the foundational approach by training on image-goal navigation and projecting language queries into multimodal semantic embedding spaces. The Target-Directed Attention Network (TDANet) \cite{10629078} advanced zero-shot capabilities through attention mechanisms learning spatial and semantic relationships using Siamese architecture designs.
Training-free approaches have emerged with Exploration with Soft Commonsense constraints (ESC) \cite{zhou2023esc}, which transfers commonsense knowledge from pre-trained models using prompt-based grounding and spatial reasoning. TriHelper \cite{zhang2024trihelper} addresses specific challenges including collision avoidance and exploration efficiency through specialized helper modules. 
Foundation model integration has enabled sophisticated zero-shot approaches. SemNav \cite{debnath2025semnav} integrates Vision Foundation Models with model-based planners for long-horizon decision making through frontier exploration, achieving promising performance in ZSON.

\subsection{LLM-Based Visual Navigation}

Large Language Models have revolutionized zero-shot navigation by providing reasoning capabilities and semantic understanding for interpreting natural language instructions \cite{wu2024voronav, kuang2024openfmnav}. OpenFMNav \cite{kuang2024openfmnav} pioneered LLM integration with vision-language models for open-set navigation, using LLMs to extract object proposals and VLMs to build Versatile Semantic Score Maps (VSSM) for exploration.
Beyond navigation, multi-agent LLM frameworks such as MALMM \cite{singh2024malmm} distribute high-level planning and low-level control across specialized LLM agents with a supervisor for zero-shot robotic manipulation. Cognitive Planning frameworks \cite{arjun2024cognitive} integrate LLMs with Large Vision-Language Models for efficient exploration strategies using 3D modular scene representations. LIM2N \cite{zu2024language} enables multimodal input processing where language and hand-drawn inputs serve as navigation constraints.
Advanced reasoning approaches include Tree-of-Thought methods in VLTNet \cite{wen2025zero} for navigation frontier selection with multi-path reasoning and backtracking capabilities. Chain-of-Thought integration in CL-CoTNav \cite{cai2025cl} provides closed-loop hierarchical reasoning with structured feedback in decision-making. World model integration in WMNav \cite{nie2025wmnav} predicts outcomes and builds memories using Curiosity Value Maps for dynamic policy configuration. NavGPT \cite{zhou2024navgpt} demonstrated early purely LLM-based navigation through zero-shot sequential action prediction with explicit high-level planning capabilities. Beyond navigation, hybrid reasoning models such as Qwen3~\cite{yang2025qwen3} can switch between thinking and non-thinking modes, and recent methods learn when to invoke deliberate reasoning through reinforcement learning~\cite{zhang2025adaptthink,tu2025learning}.

\subsection{Uncertainty Quantification in Embodied AI}

Uncertainty quantification is essential for safe and reliable robot decision-making. Classical approaches such as MC-Dropout~\cite{gal2016dropout} and Deep Ensembles~\cite{lakshminarayanan2017simple} estimate epistemic uncertainty by performing multiple stochastic forward passes or training several independent networks, respectively. While effective in controlled settings, these methods incur substantial computational overhead, making them impractical for real-time robotic deployments on resource-constrained platforms.

Conformal Prediction (CP)~\cite{angelopoulos2021gentle} has recently gained traction as a lightweight, distribution-free alternative that provides rigorous statistical coverage guarantees without retraining. In robotics, CP has been applied to safe motion planning among dynamic agents~\cite{lindemann2023safe, dixit2023adaptive}, where prediction sets over future trajectories enable conservative yet efficient collision avoidance. Ren et al.~\cite{ren2023robots} introduced KnowNo, which applies CP to LLM planners so that a robot asks for help when the prediction set contains multiple options, and later used CP for embodied question answering~\cite{ren2024explore}, where prediction set sizes decide when additional exploration is needed.

In the context of LLMs, quantifying output uncertainty remains an open challenge. Kadavath et al.~\cite{kadavath2022language} showed that large language models are reasonably well calibrated on multiple-choice questions and can estimate the probability that their own answers are correct, though such self-evaluation lacks formal statistical guarantees. Kumar et al.~\cite{kumar2023conformal} applied CP to LLM outputs in multiple-choice question answering, showing that conformal prediction sets achieve the target coverage and that their uncertainty estimates closely track prediction accuracy. However, existing work has not explored CP-based uncertainty quantification for LLM-guided robotic navigation. Our work addresses this gap by using CP to dynamically gate between fast direct inference and slow deliberative reasoning, thus providing the first principled mechanism for adaptive LLM inference in embodied navigation tasks.

\section{Method}
\label{sec:method}

\subsection{Problem Formulation}
\label{sec:problem_formulation}

We formulate the ZSON task as a Partially Observable Markov Decision Process (POMDP), defined by the tuple $\mathcal{P} = \langle \mathcal{S}, \mathcal{A}, \mathcal{T}, \mathcal{O}, \mathcal{R}, \gamma \rangle$. In this setting, an agent is initialized at a random starting pose $p_0$ within a previously unseen environment $\mathcal{E}$. The agent is assigned a semantic target goal $G$ from an open-set category vocabulary $\mathcal{V}$ (e.g., ``Bed'' or ``Sink''). Unlike traditional ObjectNav, which limits $G$ to a fixed set of training categories, ZSON requires the agent to generalize to novel objects without prior specific training.

At each discrete time step $t$, the agent receives a sensory observation $o_t$ comprising an egocentric RGB-D image pair $\langle I^{rgb}_t, I^{depth}_t \rangle$ and its current sensor pose $p_t \in SE(2)$. Based on the current observation and its historical context, the agent's policy $\pi(a_t | o_t, o_{t-1}, \dots, o_0)$ selects an action $a_t$ from a discrete action space $\mathcal{A}$. This action space consists of six primitives: Move forward ($0.25$m), Turn left ($30^{\circ}$), Turn right ($30^{\circ}$), Look up, Look down, and Stop.

The objective is to navigate to a location close to any instance of the target category $G$ (denoted as $g^*$) and execute the Stop action. An episode is considered successful if and only if two conditions are met:
\begin{enumerate}
    \item The agent issues the Stop command within a maximum episode length $T_{max}$ (e.g., 500 steps).
    \item The geodesic distance $d_{geo}(p_T, g^*)$ between the agent's stopping position $p_T$ and the nearest instance of the target object is less than a success threshold $\delta_{succ}$ (typically set to $1.0$m).
\end{enumerate}

\subsection{Overview of CovNav}

\begin{figure*}[t]
    \centering
    \includegraphics[width=1.0 \textwidth]{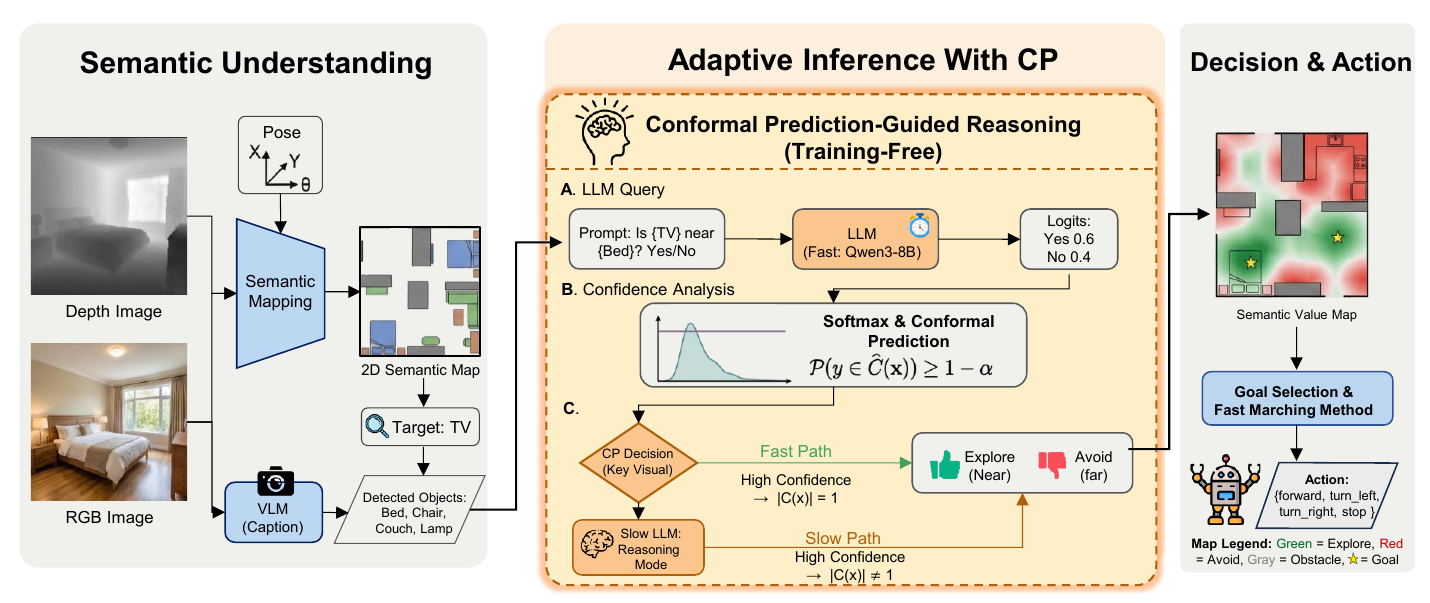}
\caption{Overview of the \textbf{CovNav} framework. The agent incrementally builds an object-based semantic map from RGB-D observations. To efficiently infer object-goal relationships, we introduce a \textbf{Fast-Slow} inference mechanism: CP dynamically evaluates the uncertainty of the local LLM operating in Fast direct inference mode. Confident predictions (singleton sets) are directly utilized, while ambiguous cases trigger the same model in Slow deliberative reasoning mode. The resulting semantic scores are fused into a global value map to guide the agent toward the most promising exploration frontiers.}
    \label{fig:framework}
\end{figure*}

Figure \ref{fig:framework} illustrates the \textbf{CovNav} framework, which synergizes perception with adaptive inference for ZSON. A \textbf{Semantic Mapping} module incrementally builds an open-vocabulary object map from RGB-D observations. To guide exploration, LLMs estimate probabilistic spatial relationships between landmarks and the target. A core innovation is the \textbf{Fast-Slow} mechanism: using Conformal Prediction (CP) to quantify uncertainty, the system first uses the local LLM in a fast direct inference mode (System 1), and activates the same model in a slow deliberative reasoning mode (System 2) only when uncertainty is high. Finally, these estimates fuse into a global value map to direct the agent toward the most promising frontiers.

\subsection{Perception: Object-wise Semantic Mapping}

The robot maintains an open-set semantic map in real-time by processing RGB-D images and positions. 
The semantic map has a $4 \times M \times M$ grid structure, where $M \times M$ represents the map resolution. The four channels consist of an obstacle channel, an exploration area channel, a target goal channel, and an item category channel. 
Specifically, the scalar values in the category channel correspond to the index of the detected object at that location.
To address the limitation of object detectors needing predefined categories, we first employ the BLIP model~\cite{li2022blip} for real-time scene understanding on single-frame RGB images. Similar to \cite{ren2024grounded}, we generate caption text for the real-time RGB image and then use NLTK~\cite{bird2006nltk} to extract key nouns from the text as prompts for the visual detection model. Note that the inference latency of both BLIP and NLTK is negligible, ensuring the pipeline operates in real-time.
We utilize the off-the-shelf segmentation model, YOLO-v7, combined with Mobile-SAM, to perform semantic segmentation on RGB images and identify common object masks. 
For objects not in the COCO~\cite{lin2014microsoft} list, we follow the VLFM~\cite{yokoyama2024vlfm} setting and use GroundingDINO to capture open-set objects outside the COCO list.
Each object mask is then mapped to a 3D semantic point cloud based on the agent's orientation and depth image. A 2D semantic map is created by projecting the 3D point cloud from top to bottom onto a 2D plane, mapping all point clouds containing semantic information to obstacle channels and their respective object type channels. Point clouds near the floor are mapped to the explored channel.

\subsection{Exploration with LLM Guidance}

\subsubsection{Object Relationship}

To explore unseen environments efficiently, the agent must infer the potential location of a not-yet-observed target goal based on observable landmarks. This inference relies on \textbf{Object Relationships}, i.e., the probabilistic spatial co-occurrences intrinsic to indoor layouts. For instance, a ``Nightstand'' strongly implies a nearby ``Bed''. Since these commonsense spatial relationships are fundamentally static and do not change across scenes, querying an LLM online for every object pair is computationally inefficient and unnecessary. To address this, we propose a hybrid inference mechanism that combines offline pre-computation for efficiency with online adaptive inference for open-set generalization.

\paragraph{Static Knowledge Memory.} For common object categories (e.g., COCO~\cite{lin2014microsoft} classes) that frequently appear in indoor environments, we pre-compute an $N \times N$ \textit{Object Relationship Matrix} $\mathcal{M}$ offline using a powerful Large Foundation Model (i.e., GPT-5). Specifically, for each pair of object categories $(c_i, c_j)$, we query GPT-5 to estimate the conditional probability $P(\text{Target}=c_j|\text{Landmark}=c_i)$ based on its vast commonsense knowledge base. These probabilities are stored in $\mathcal{M}$, allowing for $O(1)$ constant-time lookups during navigation. This ``Memory'' acts as a fast-access buffer for static commonsense, effectively reducing the latency for known object pairs.

\paragraph{Online Adaptive Inference.} For open-set objects encountered in the wild that are not covered by the pre-computed memory (i.e., $(o_{obs}, o_{goal}) \notin \mathcal{M}$), we employ a local lightweight LLM (Qwen3-8B) to perform online inference. The query is structured as a discriminative task:

\begin{mdframed}[backgroundcolor=gray!15, linewidth=0pt]
\textit{``Consider the common room layout. Is it highly likely to find a \{goal\_name\} close to the \{landmark\} in the same room? Answer with `Yes' or `No'.''}
\end{mdframed}
To obtain a continuous metric of semantic relationship, we perform a single forward pass and extract the logits for ``Yes'' ($z_{\mathrm{Yes}}$) and ``No'' ($z_{\mathrm{No}}$). The normalized confidence score $c \in [0,1]$ is computed via temperature ($T$) scaling:
\begin{equation}
    c = \frac{\exp(z_{\mathrm{Yes}}/T)}{\exp(z_{\mathrm{Yes}}/T) + \exp(z_{\mathrm{No}}/T)}
\end{equation}
This continuous score represents the estimated likelihood of the target's presence. Instead of a binary decision, it provides a dense, probabilistic metric reflecting the strength of the semantic relationship between the landmark and the target goal. However, because a single fast forward pass from a small local model can be unreliable on ambiguous object pairs, we further employ CP to quantify uncertainty and trigger a slow deliberative reasoning mode when necessary.

\subsubsection{Conformal Prediction for Uncertainty Quantification}

In the context of object goal navigation, relying solely on point predictions from an LLM (e.g., the token with the highest probability) carries the risk of confident hallucinations, especially when the spatial relationship is ambiguous (e.g., distinguishing whether a ``Chair'' implies a nearby ``Table'' or ``Toilet''). To mitigate this, we employ CP~\cite{angelopoulos2021gentle}, a distribution-free uncertainty quantification framework that transforms the model's raw confidence scores into a \textit{prediction set} with a rigorous statistical guarantee; we refer the reader to~\cite{angelopoulos2021gentle, ren2024explore} for a thorough exposition.

Formally, let $x \in \mathcal{X}$ denote the input query describing the relationship between an observed landmark and the target goal, and let $y \in \mathcal{Y} = \{\text{`Yes'}, \text{`No'}\}$ be the true label indicating whether they are spatially co-located. We assume access to a small calibration dataset $\mathcal{D}_{\text{cal}} = \{(x_i, y_i)\}_{i=1}^N$ drawn exchangeably from the same distribution as the test data. The central guarantee is that, at inference, given a new i.i.d. sample $x_{\text{test}}$ with unknown true label $y_{\text{test}}$, CP generates a prediction set $\mathcal{C}(x_{\text{test}}) \subseteq \mathcal{Y}$ that is guaranteed to contain the true label with a user-specified probability.

The construction of CP proceeds as follows. First, we define a \textit{nonconformity score} $s(x, y) = 1 - \hat{P}(y|x)$ that quantifies how ``atypical'' a label $y$ is for input $x$, where $\hat{P}$ is the model's softmax probability. Intuitively, a higher nonconformity score indicates lower model confidence in the correct answer. We then evaluate this score on all calibration samples to obtain $\{s_i = s(x_i, y_i)\}_{i=1}^N$. The key step is to compute a \textit{conformal threshold} $\hat{q}$ as the $\lceil (N+1)(1-\alpha) \rceil / N$ empirical quantile of these calibration scores, where $\alpha \in (0,1)$ is a user-specified error rate (e.g., $\alpha=0.1$ for 90\% coverage).

For a new test input $x_{\text{test}}$, the prediction set is constructed by including all labels whose nonconformity score falls below the threshold, effectively retaining all plausible answers:
\begin{equation}
\label{eq:prediction_set}
\mathcal{C}(x_{\text{test}}) = \{y \in \mathcal{Y} \mid \hat{P}(y|x_{\text{test}}) \geq 1 - \hat{q}\}.
\end{equation}
The generated prediction set ensures that the following coverage guarantee holds~\cite{angelopoulos2021gentle}:
\begin{equation}
\label{eq:coverage_guarantee}
  \mathbb{P}(y_{\text{test}} \in \mathcal{C}(x_{\text{test}})) \geq 1 - \alpha.
\end{equation}
The intuition behind this guarantee is as follows: by choosing $\hat{q}$ as the $(1-\alpha)$-quantile of the calibration nonconformity scores, we ensure that at most an $\alpha$ fraction of calibration samples have a score exceeding $\hat{q}$. Since the test sample is exchangeable with the calibration data, it will similarly have a score $\leq \hat{q}$ with probability at least $1-\alpha$. By construction, we include all such labels in $\mathcal{C}(x_{\text{test}})$, thereby ensuring the coverage guarantee.

Crucially, the size of the prediction set $|\mathcal{C}(x_{\text{test}})|$ serves as a rigorous indicator of the model's uncertainty. We leverage this cardinality as a dynamic switching signal for our Fast-Slow adaptive inference mechanism, allowing the system to distinguish between confident and ambiguous states, as detailed below.

\subsubsection{Fast-Slow Inference Switching with CP}

Next, we describe how the prediction set size $|\mathcal{C}(x)|$ serves as a switching signal between the fast direct inference and slow deliberative reasoning modes, mirroring the human cognitive duality of System 1 (fast, intuitive) and System 2 (slow, analytical)~\cite{kahneman2011thinking}. 

Each time the robot observes a new object, it first invokes the local LLM in fast direct inference mode (System 1) to assess the likelihood of finding the target goal near this object. We use the query formulation defined in Section~\ref{sec:method}-D.1 and extract the logits for `Yes' and `No' to construct the prediction set $\mathcal{C}(x)$ via Equation~\ref{eq:prediction_set}. The decision logic is as follows:

    \textbf{Singleton Set ($|\mathcal{C}(x)|=1$):} The prediction set contains exactly one label (either \{`Yes'\} or \{`No'\}). This indicates that the fast direct inference mode is confident in its judgment. In this case, we rely directly on the fast-mode output and use the softmax probability $c$ (corresponding to the ``Yes'' token) as the semantic value for the observed object.

    \textbf{Empty or Doubleton Set ($|\mathcal{C}(x)| \neq 1$):} The prediction set is either empty ($\emptyset$) or contains both labels (\{`Yes', `No'\}). An empty set suggests that neither answer meets the required confidence level (often due to high anomaly), while a doubleton suggests that both answers are plausible under the statistical guarantee (ambiguity). Both cases represent high uncertainty where fast direct inference fails to provide a unique answer. Consequently, the system re-invokes the same model in slow deliberative reasoning mode (System 2) to perform step-by-step reasoning, and the result from this deeper analysis is used as the target value.

When the slow mode is triggered, the same Qwen3-8B model is invoked with the \texttt{enable\_thinking} flag activated. This enables the model to generate an internal chain-of-thought reasoning trace before producing the final answer. The reasoning process typically involves: (1) identifying the functional category of the observed landmark (e.g., ``a couch is typically found in a living room''), (2) inferring the room type most likely associated with the target goal, and (3) evaluating cross-room spatial plausibility. This deliberative process yields a recalibrated confidence score that replaces the original uncertain estimate.

\subsubsection{Collection of Calibration Set}

To rigorously quantify uncertainty using CP, we require a calibration set $\mathcal{D}_{\text{cal}} = \{(x_i, y_i)\}_{i=1}^N$ that accurately reflects typical indoor object arrangements. Instead of relying on manual annotation, which is unscalable, or real-world scanning, which is data-sparse, we propose a scalable, automated pipeline to construct $\mathcal{D}_{\text{cal}}$ by distilling commonsense knowledge from LLMs and datasets. Our pipeline consists of three steps: Scene Generation, Object Filtering, and Pairwise Labeling.

    \textbf{Scene Description Generation.} 
    To ensure calibration data reflects realistic indoor object distributions, we randomly sample viewpoints from the Matterport3D training split across four primary room types: \textit{bedroom, living room, kitchen,} and \textit{bathroom}. For each sampled viewpoint, we utilize GPT-5 as a vision-language model to describe the RGB image, extracting a list of visible objects and the corresponding room category. For example, a bedroom viewpoint might yield \textit{``bed, nightstand, lamp, wardrobe, ...''}. We collect 50 viewpoints per room type to ensure diversity. All generated object lists are manually screened and verified for rationality.

    \begin{figure}[t]
    \centering
    \includegraphics[width=1.0\linewidth]{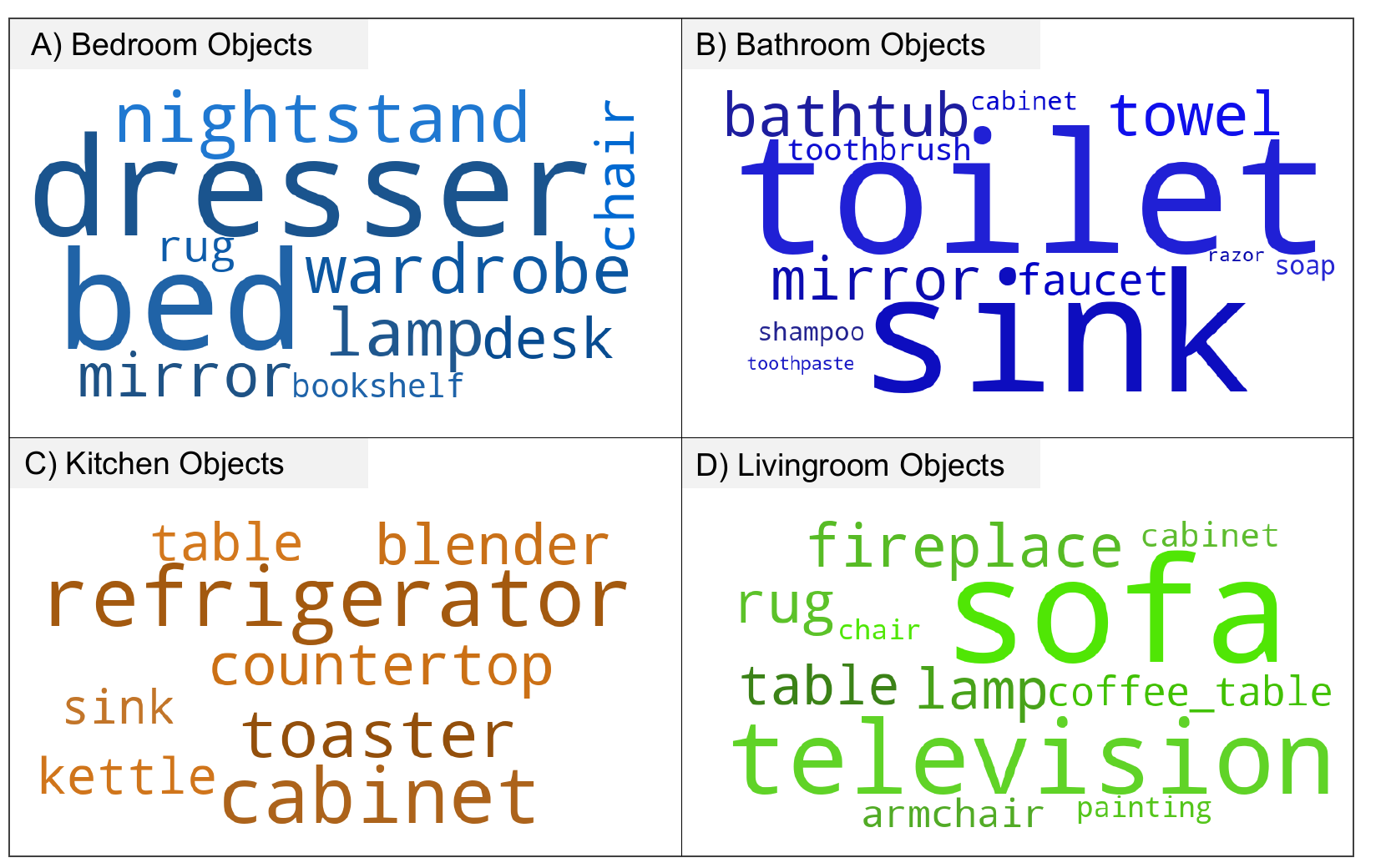}
    \caption{Word cloud visualization of object frequencies across four room types. Larger font sizes indicate higher frequency of occurrence, highlighting the characteristic components validation.}
    \label{fig:wordcloud}
\end{figure}
    \textbf{Frequency-based Object Filtering.} 
    Raw descriptions may contain rare artifacts or hallucinations. To extract robust characteristic objects, we perform frequency analysis on the generated corpus. We calculate the occurrence count of each unique object within specific room categories.
    As shown in Figure~\ref{fig:wordcloud}, objects cluster naturally by room function. We apply a filtering threshold $\tau_{freq} = 20\%$ (appearing in $>$10/50 descriptions) to identifying \textit{Characteristic Components} (e.g., \textit{sofa} for living room, \textit{oven} for kitchen). This step effectively filters out noise while retaining essential spatial elements.

    \textbf{Automated Set Construction.} 
    Finally, we synthesize labeled query pairs $(x, y)$ to form the calibration set. The labeling logic is grounded in the functional coherence of indoor spaces. \textbf{Positive samples} ($y=$ ``Yes'') are generated from pairwise permutations of characteristic components belonging to the \textit{same} room category (e.g., Bed $\leftrightarrow$ Nightstand), representing high-probability spatial assignments. Conversely, \textbf{negative samples} ($y=$ ``No'') are drawn from pairwise permutations of components across \textit{distinct} room categories (e.g., Bed $\leftrightarrow$ Toilet), representing implausible spatial proximities.
    This structured approach yields a clean, balanced, and semantically valid dataset $\mathcal{D}_{\text{cal}}$ for CP calibration with human feedback.

\subsubsection{Global Decision with Semantic Exploration}

To guide the agent effectively, we construct a Global Semantic Value Map by aggregating the semantic contributions of all observed objects. This process operates in three stages:

\noindent\textbf{Individual Influence Field.} For each discovered object $o_i$ at location $\mathbf{l}_i$, we obtain a base semantic score $v_i$ from the LLM (positive for target-associated landmarks, negative otherwise). Rather than treating this as a point value, we model the object's influence as a spatial field that decays exponentially with distance. The value at any map coordinate $p$ induced by object $o_i$ is given by:
\begin{equation}
    V_i(p) = v_i \cdot \exp\left(-\frac{\|p - \mathbf{l}_i\|}{\lambda}\right)
\end{equation}
where $\lambda$ is a decay constant controlling the spatial extent of influence. We set $\lambda = 1.0$m, which corresponds to typical room-scale spatial relationships: an object's semantic signal remains significant within its immediate vicinity but diminishes beyond adjacent rooms.

\noindent\textbf{Global Map Fusion.} We obtain the Global Semantic Value Map $M_{sem}$ by superimposing these individual influence fields. By summing the values from all objects, $M_{sem}(p) = \sum_i V_i(p)$, overlapping fields from spatially clustered positive landmarks mutually reinforce each other, creating high-value regions that attract the agent. Conversely, negative landmarks create repulsive zones. 

\noindent\textbf{Frontier Selection.} To balance goal-seeking with exploration, we integrate a Historical Value Map $M_{hist}$. The agent's historical trajectory is projected onto the map with similar exponential decay to penalize revisiting explored areas. The final decision is made by fusing these maps: $M_{total} = M_{sem} + M_{hist}$. Frontier nodes are defined as the boundary cells between explored and unexplored regions on the obstacle map. The agent selects the frontier node with the highest cumulative value on $M_{total}$ as the next mid-term goal, effectively prioritizing unexplored areas that are most semantically promising.

\subsection{Local Planning}

The navigation process operates in a hierarchical manner. At fixed time intervals $\Delta t = 25$ steps, the robot invokes the global decision module described above to re-evaluate the semantic value map and select the next mid-term goal from the frontier nodes. This periodic replanning allows the agent to incorporate newly discovered objects and dynamically adjust its exploration strategy as the environment unfolds. We empirically find that this interval balances responsiveness to new observations with computational overhead.

Between successive global decisions, a local planner guides the robot toward the current mid-term goal. Given the robot's current position and the selected target point, the Fast Marching Method \cite{sethian1996fast} is employed to compute the shortest feasible path on the obstacle map in real-time, generating a sequence of waypoints. The path is computed on the obstacle channel of the semantic map, ensuring that the planned trajectory avoids known obstacles while respecting the map resolution of $0.05$m per cell. To track this path, the agent first checks the proximity to the goal; if the distance is within $0.25$m, the Stop action is triggered. Otherwise, it computes the angular deviation $\theta_{dev}$ between its current heading and the direction to the immediate next waypoint. Discrete actions are then deterministically derived from this deviation: if $|\theta_{dev}| > 15^{\circ}$, a rotational action (Turn Left if $\theta_{dev} > 0$, Turn Right otherwise) is selected to align the agent; otherwise, a Move Forward action is executed.

\section{Experiments}
\label{sec:experiments}

\begin{figure*}[t]
    \centering
    \includegraphics[width=0.9\linewidth]{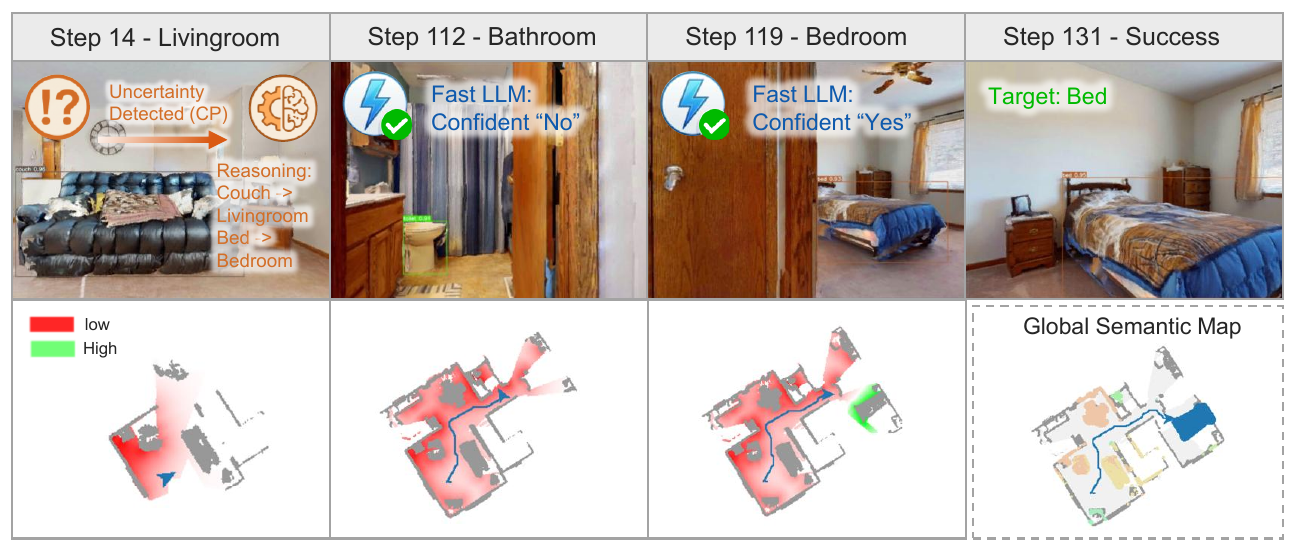}
    \caption{Visualization of the adaptive inference process (Target: Bed). At Step 14, the agent encounters semantic ambiguity while observing a couch. The CP module detects high uncertainty, triggering the Slow Mode to perform chain-of-thought reasoning (inferring ``Living Room'' from ``Couch'' and ``Bedroom'' from ``Bed''), thereby guiding the agent to the correct room. At Step 112, the Fast Mode effectively filters out irrelevant areas (Bathroom) with high confidence, leading to a successful identification of the target at Step 119. The bottom row illustrates the real-time evolution of the Global Semantic Value Map (visualized as a red-green heatmap) and the agent's trajectory.}
    \label{fig:visualization}
\end{figure*}

\subsection{Simulation Setting}
\label{subsec:exp_setting}
{\bf Simulation Platform.} We conduct all simulated experiments on the Habitat simulator~\cite{savva2019habitat}, a high-performance 3D embodied AI platform that supports realistic indoor scene rendering.
The simulator provides the navigation agent with RGB-D observations at a resolution of $640 \times 480$. The Horizontal Field of View (HFOV) is $79^{\circ}$ for both RGB and depth sensors, with a valid depth range of $0.5$m to $5$m.
It also offers rich sensor information, real-time agent states, and precise physics updates following each action.
The action space includes six discrete actions: move forward ($0.25$m), turn left ($30^{\circ}$), turn right ($30^{\circ}$), look up, look down, and stop. 
We follow the task setting of the Habitat Challenge 2022~\cite{habitatchallenge2022} and VLFM~\cite{yokoyama2024vlfm}.
Each episode has a maximum time budget of $500$ steps.

{\bf Datasets.} We evaluate our method on three mainstream indoor scene datasets provided by the Habitat simulator: Gibson, HM3D v1, and HM3D v2. These datasets feature a wide variety of complex and realistic indoor scenes, composed of colored 3D scans. Specifically, Gibson provides $5$ different scenes with a total of $1,000$ episodes. HM3D v1 and v2 provide $20$ and $36$ scenes, offering $2,000$ and $1,000$ episodes, respectively. The target objects for all episodes belong to $6$ common indoor categories.

{\bf Metrics.} We adopt Success Rate (SR) and Success weighted by Path Length (SPL) as our primary evaluation metrics \cite{anderson2018evaluation}. We conduct $N$ test episodes. In each episode $i$, let $l_i$ be the shortest-path distance from the agent's starting position to the goal, and $p_i$ be the length of the path actually taken by the agent. Let $S_i$ be a binary indicator of success in episode $i$. Then, SR and SPL are calculated as follows:
\begin{equation}
\text{SR} = \frac{1}{N} \sum_{i=1}^{N} S_i
\end{equation}
\begin{equation}
\text{SPL} = \frac{1}{N} \sum_{i=1}^{N} S_i \frac{l_i}{\max(p_i, l_i)}
\end{equation}
SR measures the effectiveness of the agent, while SPL jointly evaluates success and navigation efficiency.

{\bf LLM Configuration.} 
We employ a hybrid knowledge retrieval system combined with a fast-slow inference mechanism. The configuration details are as follows:

\begin{itemize}[leftmargin=10pt, itemsep=-2pt]
\item \textbf{Static Memory.} We utilize GPT-5 to pre-compute the probabilistic object relationship matrix for the 80 COCO categories, serving as a static commonsense memory for $O(1)$ retrieval.
\item \textbf{Online Inference.} For open-set objects, we employ Qwen$3$-$8$B as the local inference engine. The \textbf{Fast Mode} uses the base model for rapid direct inference, while the \textbf{Slow Mode} activates the \texttt{enable\_thinking} flag for deliberative reasoning when high uncertainty is detected.
\end{itemize}
We set the temperature parameter $T=10$ for scaling the logits. The conformal prediction calibration is performed with a user-specified error rate $\alpha=0.5$ on a calibration set containing $1{,}200$ object-pair relationship queries constructed from four typical room categories.

{\bf Perception Module.} We adopt YOLOv7~\cite{wang2023yolov7} for common object detection, and GroundedSAM~\cite{ren2024grounded} processing RGB frames to extract semantic information. The detection confidence threshold is set to $0.5$ to balance precision and recall. The semantic map has a spatial resolution of $0.05$m per grid cell with dimensions of $480 \times 480$ cells, covering a $24$m $\times$ $24$m area.

{\bf Hardware Resources.} All simulated experiments, including both fast and slow mode inference, are conducted on a workstation equipped with an NVIDIA RTX 4090D GPU ($24$GB VRAM), AMD Ryzen $9$ $7950$X CPU, and $64$GB RAM.

\begin{table*}[htbp]
    \caption{\textbf{Comparison with ZSON Baselines.}
   }
  
  \vskip 0.05in
  
    \centering
    \resizebox{.9\textwidth}{!}{%
    \begin{tabular}{lccccccccc}
      
      \toprule
      \multirow{2}{*}{Method} & \multirow{2}{*}{Planner} &\multirow{2}{*}{Training-free} & \multirow{2}{*}{Core Model} & \multicolumn{2}{c}{Gibson} &\multicolumn{2}{c}{HM3Dv1} & \multicolumn{2}{c}{HM3Dv2}  \\
       \cmidrule(l{1pt}r{1pt}){5-6}\cmidrule(l{1pt}r{1pt}){7-8}\cmidrule(l{1pt}r{1pt}){9-10}
      &&&& Success$\uparrow$ & SPL$\uparrow$ & Success$\uparrow$ & SPL$\uparrow$ & Success$\uparrow$ & SPL$\uparrow$ \\
      \midrule
      L3MVN & \multirow{5}{*}{Semantic} & X &LLM& 76.9  & 38.8 & 38.4 & 19.4  & 36.3& 15.7 \\
       VLFM && $\checkmark$ & VLM & 81.8 & 46.5 & 50.9 & 23.6 & 56.9 & 27.5  \\
       GAMap && $\checkmark$ & VLM & \textbf{85.7} & \textbf{55.5} & 53.1 & 26.0 & - & -  \\
       ApexNav && $\checkmark$ &   VLM & - & - & 59.6 & 33.0 & 76.2 & 38.0   \\
       \textbf{CovNav (Ours)} && $\checkmark$ & LLM & 82.4  & 48.6  & \textbf{60.4} & \textbf{33.1} & \textbf{77.9} & \textbf{40.2}  \\

      \bottomrule
  
    \end{tabular}%
      }
    \label{tab:main_results}
  \end{table*}

\subsection{Simulation Comparison}
\label{subsec:sim_comparison}

We compare CovNav with state-of-the-art methods on Gibson and HM3D datasets. The quantitative results are presented in Table~\ref{tab:main_results}.

\textbf{Baselines.} We compare CovNav against state-of-the-art ZSON methods: 
(1) \textbf{L3MVN}~\cite{yu2023l3mvn}, which uses LLMs to build a semantic value map for finding objects;
(2) \textbf{VLFM}~\cite{yokoyama2024vlfm}, which builds a value map using BLIP-2 to guide exploration;
(3) \textbf{GAMap}~\cite{huang2024gamap}, which guides exploration with multi-scale geometric and affordance attributes of the target object; 
(4) \textbf{ApexNav}~\cite{zhang2025apexnav}, which adaptively switches between semantic-driven and geometry-driven exploration and employs target-centric semantic fusion to reduce false detections.

\textbf{Analysis.}
The results in Table~\ref{tab:main_results} reveal a compelling trend: while baseline methods like GAMap excel in simpler environments (Gibson), their performance degrades significantly in complex, large-scale scenes (HM3D). Specifically, GAMap achieves a leading 85.7\% SR on Gibson but drops to 53.1\% on HM3D v1. This suggests that relying solely on geometric frontiers or uncalibrated semantic maps is insufficient when spatial ambiguity increases. In contrast, CovNav demonstrates superior generalization. On the challenging HM3D v1 and v2 benchmarks, it outperforms the runner-up ApexNav by margins of 0.8\% and 1.7\% respectively, while maintaining competitive efficiency. The key differentiator is CovNav's ability to \textit{recognize when it doesn't know}. By adaptively escalating from fast direct inference to slow deliberative reasoning in ambiguous scenarios, CovNav approaches the performance of the Slow-Only variant without paying its full computational cost, which is particularly beneficial for complex HM3D environments where such ambiguity is more prevalent. The consistently high SPL scores further indicate that our CP-guided exploration reduces ``random walking'' behavior, leading to more direct paths to the goal.

\subsection{Ablation Study}
\label{subsec:ablation}

We conduct extensive ablation studies to validate the effectiveness of the proposed fast-slow inference mechanism and the impact of different system components. For computational efficiency, we randomly sample $100$ episodes from each dataset (Gibson, HM3D v1, and HM3D v2) to perform the following ablation experiments. In these ablations, SR and SPL are first computed for each target category and then averaged over the six categories.
\subsubsection{Fast-Slow Inference Mechanism}
To evaluate the efficacy of our adaptive inference approach, we compare CovNav against three variants: 
(1) \textbf{Fast-Only (System 1)}: Uses only the 8B model in fast direct inference mode; 
(2) \textbf{Slow-Only (System 2)}: Consistently employs the same 8B model in slow deliberative reasoning mode; 
(3) \textbf{Random Switch}: Randomly switches between fast and slow modes.
The results in Table~\ref{tab:ablation_fast_slow} highlight the trade-off between performance and efficiency. The time of Random Switch and CovNav is estimated from the module-level latencies reported in Table~\ref{tab:efficiency}.

The performance gap between Fast-Only and Slow-Only (e.g., $51.5\%$ vs. $64.2\%$ on HM3D v1) empirically validates the limitation of relying solely on fast direct inference from a small local model: single-pass inference lacks the deliberative capacity to handle complex spatial relations consistently. While the Slow-Only variant addresses this, its $3.40$s latency is operationally infeasible. CovNav effectively breaks this ``accuracy-efficiency'' curse. It recovers nearly the entire performance gain of the slow mode (achieving $62.8\%$ SR, within $1.4\%$ of the upper bound) while keeping latency low ($0.72$s). Crucially, the $12.4\%$ slow-mode activation rate reveals that \textit{most} navigation decisions are routine; however, the few critical decisions that require slow deliberative reasoning are determinant for success. CovNav's ``strategic slowness'' ensures resources are spent exactly on these high-leverage moments, rather than being wasted on trivial steps or withheld during critical ambiguities requiring deliberation.

\begin{table}[htbp]
\centering
\caption{\textbf{Ablation Study of Fast-Slow Inference Mechanism.} Comparison of navigation performance and inference efficiency on Gibson and HM3D datasets.}
\label{tab:ablation_fast_slow}
\resizebox{\columnwidth}{!}{%
\begin{tabular}{lccccc}
\toprule
\multirow{2}{*}{Method} & Gibson & HM3D v1 & HM3D v2 & \multicolumn{2}{c}{Efficiency} \\
\cmidrule(lr){2-2} \cmidrule(lr){3-3} \cmidrule(lr){4-4} \cmidrule(lr){5-6}
& SR $\uparrow$ / SPL $\uparrow$ & SR $\uparrow$ / SPL $\uparrow$ & SR $\uparrow$ / SPL $\uparrow$ & Time (s) & Call (\%) \\
\midrule
Fast-Only (System 1) & 72.5 / 38.2 & 51.5 / 25.8 & 69.8 / 34.2 & \textbf{0.63} & - \\
Slow-Only (System 2) & \textbf{83.1} / \textbf{49.0} & \textbf{64.2} / 31.5 & \textbf{82.5} / \textbf{44.8} & 3.40 & 100 \\
Random Switch & 76.8 / 42.5 & 57.1 / 29.8 & 75.4 / 38.1 & 1.25 & 53.0 \\
\textbf{CovNav (Adaptive)} & 79.5 / 43.7 & 62.8 / \textbf{32.8} & 78.5 / 39.9 & 0.72 & 12.4 \\
\bottomrule
\end{tabular}%
}
\end{table}

\subsubsection{Impact of Conformal Prediction Parameter $\alpha$}
The parameter $\alpha$ in Conformal Prediction controls the tolerance for uncertainty. We analyze how varying $\alpha$ affects the switching rate between fast and slow modes and the overall navigation performance. Table~\ref{tab:ablation_alpha} shows the sensitivity analysis.

\begin{table}[htbp]
\centering
\caption{\textbf{Impact of Conformal Prediction Parameter $\alpha$.}\\ Evaluating the trade-off between switching rate and navigation performance across datasets.}
\label{tab:ablation_alpha}
\resizebox{\columnwidth}{!}{%
\begin{tabular}{ccccc}
\toprule
\multirow{2}{*}{$\alpha$} & \multirow{2}{*}{Switch (\%)} & Gibson & HM3D v1 & HM3D v2 \\
\cmidrule(lr){3-3} \cmidrule(lr){4-4} \cmidrule(lr){5-5}
& & SR $\uparrow$ / SPL $\uparrow$ & SR $\uparrow$ / SPL $\uparrow$ & SR $\uparrow$ / SPL $\uparrow$ \\
\midrule
0.8 & 22.4 & \textbf{82.9} / \textbf{48.8} & 61.5 / 31.9 &   \textbf{79.2} / 39.1\\
0.5 (Ours) & 12.4 &  79.5 / 43.7 & \textbf{62.8} / \textbf{32.8} &  78.5 / \textbf{39.9}  \\
0.2 & 8.2 & 80.1 / 46.5 & 59.4 / 30.1 & 75.8 / 37.5 \\
\bottomrule
\end{tabular}%
}
\end{table}

The CP parameter $\alpha$ serves as a tunable ``risk tolerance'' knob. Raising $\alpha$ to $0.8$ creates a conservative agent that triggers the slow deliberative mode more frequently ($22.4\%$), which explains its highest SR on Gibson ($82.9\%$) and HM3D v2 ($79.2\%$), as the additional deliberation overhead pays off in these environments. However, in the most challenging HM3D v1 scenes, excessive switching introduces decision noise, causing SR to fall below our default setting ($61.5\%$ vs. $62.8\%$). Conversely, a low $\alpha$ ($0.2$) creates an overconfident agent that rarely invokes the slow mode ($8.2\%$), failing to trigger it when genuinely needed. This leads to consistent performance degradation across all datasets (e.g., $75.8\%$ on HM3D v2), approaching the ``Fast-Only'' baseline. Our choice of $\alpha=0.5$ achieves an effective trade-off between accuracy and cost, identifying a ``sweet spot'' where the system is sufficiently self-aware to catch errors without becoming paralyzed by computational overhead.

\subsubsection{Mode Switching without Uncertainty}
We investigate the performance when using fixed probability thresholds for fast-slow mode switching instead of the adaptive CP-based approach. Table~\ref{tab:ablation_threshold} presents the results for different fixed thresholds across datasets.

\begin{table}[htbp]
\centering
\caption{\textbf{Performance with Fixed Decision Thresholds.}\\ Comparison of decision-making without uncertainty quantification.}
\label{tab:ablation_threshold}
\resizebox{\columnwidth}{!}{%
\begin{tabular}{ccccc}
\toprule
\multirow{2}{*}{Threshold} & \multirow{2}{*}{Switch (\%)} & Gibson & HM3D v1 & HM3D v2 \\
\cmidrule(lr){3-3} \cmidrule(lr){4-4} \cmidrule(lr){5-5}
& & SR $\uparrow$ / SPL $\uparrow$ & SR $\uparrow$ / SPL $\uparrow$ & SR $\uparrow$ / SPL $\uparrow$ \\
\midrule
0.8 & 42.7 & \textbf{82.0} / \textbf{45.7} & 54.2 / 26.5 & \textbf{80.4} / \textbf{41.6} \\
0.5 & 10.1 & 77.8 / 44.5 & 60.9 / 31.4 & 77.1 / 38.2 \\
0.2 & 8.2 & 74.2 / 39.8 & 57.4 / 28.6 & 74.6 / 36.4 \\
\textbf{CovNav (CP)} & 12.4 & 79.5 / 43.7 & \textbf{62.8} / \textbf{32.8} & 78.5 / 39.9 \\
\bottomrule
\end{tabular}%
}
\end{table}

Increasing the fixed probability threshold imposes a stricter safety requirement, causing the agent to trigger the slow deliberative mode more frequently. For instance, a conservative threshold of $0.8$ leads to a $42.7\%$ switching rate. While this aggressive strategy yields high success on the simpler Gibson dataset ($82.0\%$), it fails to generalize to complex scenes like HM3D v1 (dropping to $54.2\%$), barely surpassing the Fast-Only baseline. This limitation stems from the uncalibrated nature of raw logits: the model can be ``wrong and confident'' (hallucination), bypassing the threshold check. In contrast, CP acts as a statistical rectifier. By calibrating against a held-out set, CP ensures the uncertainty metric reflects \textit{true empirical correctness}. Consequently, CovNav achieves superior performance on HM3D v1 ($62.8\%$) with significantly lower computational cost ($12.4\%$ switching), proving that \textit{correctly quantifying uncertainty} is more effective than merely increasing the safety threshold.

\subsubsection{System Components}
We further decompose the system to validate the contribution of each module, including the semantic fusion strategy, confidence decay, and open-set perception, as follows.

\begin{table}[htbp]
\centering
\caption{\textbf{Ablation of System Components.} Analyzing the contribution of different modules.}
\label{tab:ablation_components}
\resizebox{\columnwidth}{!}{%
\begin{tabular}{lccc}
\toprule
\multirow{2}{*}{Method} & Gibson & HM3D v1 & HM3D v2 \\
\cmidrule(lr){2-2} \cmidrule(lr){3-3} \cmidrule(lr){4-4}
& SR $\uparrow$ / SPL $\uparrow$ & SR $\uparrow$ / SPL $\uparrow$ & SR $\uparrow$ / SPL $\uparrow$ \\
\midrule
w/o CP-calibrated Value & 52.5 / 25.2 & 41.2 / 21.9 & 66.8 / 28.5 \\
w/o Confidence Decay & 80.2 / 46.8 & 51.3 / 25.8 & 65.4 / 31.2 \\
w/o Open-set Perception & 68.4 / 35.1 & 54.6 / 27.5 & 69.5 / 34.1 \\
\textbf{CovNav (Ours)} &  79.5 / 43.7 & 62.8 / 32.8 & 78.5 / 39.9 \\
\bottomrule
\end{tabular}%
}
\end{table}

The ablation results confirm that CovNav's performance is not due to a single component but the synergy of its parts. First, the \textbf{CP-calibrated Value} is critical for efficient search. Removing this component (w/o CP-calibrated Value) effectively degrades the system to \textbf{traditional frontier-based exploration}, where the robot explores space based on geometric frontiers without semantic prioritization. Consequently, performance drops significantly across all datasets (e.g., $62.8\% \rightarrow 41.2\%$ on HM3D v1), confirming that semantic guidance is indispensable. \textbf{Confidence decay} is equally vital; without it, the map becomes ``stale'' and dominated by historical errors, dropping SR to $51.3\%$ on HM3D. Finally, \textbf{Open-set perception} is the foundational bedrock; reverting to closed-set detection artificially limits the agent's understanding of context, causing a drastic performance collapse ($54.6\%$). Thus, CovNav succeeds by combining rich open-set data (Perception) with fresh, reliable interpretation (Decay + CP).

\subsection{Efficiency and Runtime Analysis}
\label{subsec:efficiency}

We analyze the computational resource consumption of the proposed method, focusing on model size, memory usage, and latency. Table~\ref{tab:efficiency} provides a detailed breakdown.

\begin{table}[htbp]
\centering
\caption{\textbf{Computational Cost Analysis.} Comparison of resource usage and latency across modules.}
\label{tab:efficiency}
\resizebox{\columnwidth}{!}{%
\begin{tabular}{lccc}
\toprule
Module & Model Size & VRAM (GB) & Latency (ms) \\
\midrule
Perception (BLIP\&YOLO) & 125M & \textbf{4.2} & \textbf{45} \\
Fast Mode (Qwen3-8B-FP8) & 8B & 8.0 & 180 \\
GroundedSAM & 172M & 3.8 & 300 \\
Slow Mode (Thinking Enabled) & 8B & 8.0 & 2400 \\
\textbf{CovNav (Avg)} & - & 22.6 & 680 \\ 
\bottomrule
\end{tabular}%
}
\end{table}

The efficiency analysis underscores the practical advantage of CovNav. While the thinking-enabled slow mode provides higher-quality deliberative inference, its latency ($2400$ms) is prohibitive for real-time control loops. By leveraging the Fast Mode ($180$ms) for the majority of queries, CovNav reduces the average inference latency to $680$ms, a $3.5$ times speedup over the slow-only approach. This latency is well within the acceptable range for mobile robot navigation, allowing for smooth and responsive behavior. Furthermore, the total memory footprint is $22.6$GB, fitting comfortably within the memory constraints of consumer-grade GPUs (e.g., RTX 3090/4090) commonly found on high-end robotic workstations.

\subsection{Calibration Analysis}
\label{subsec:calibration}

\begin{figure}[t]
    \centering
    \includegraphics[width=1.0\linewidth]{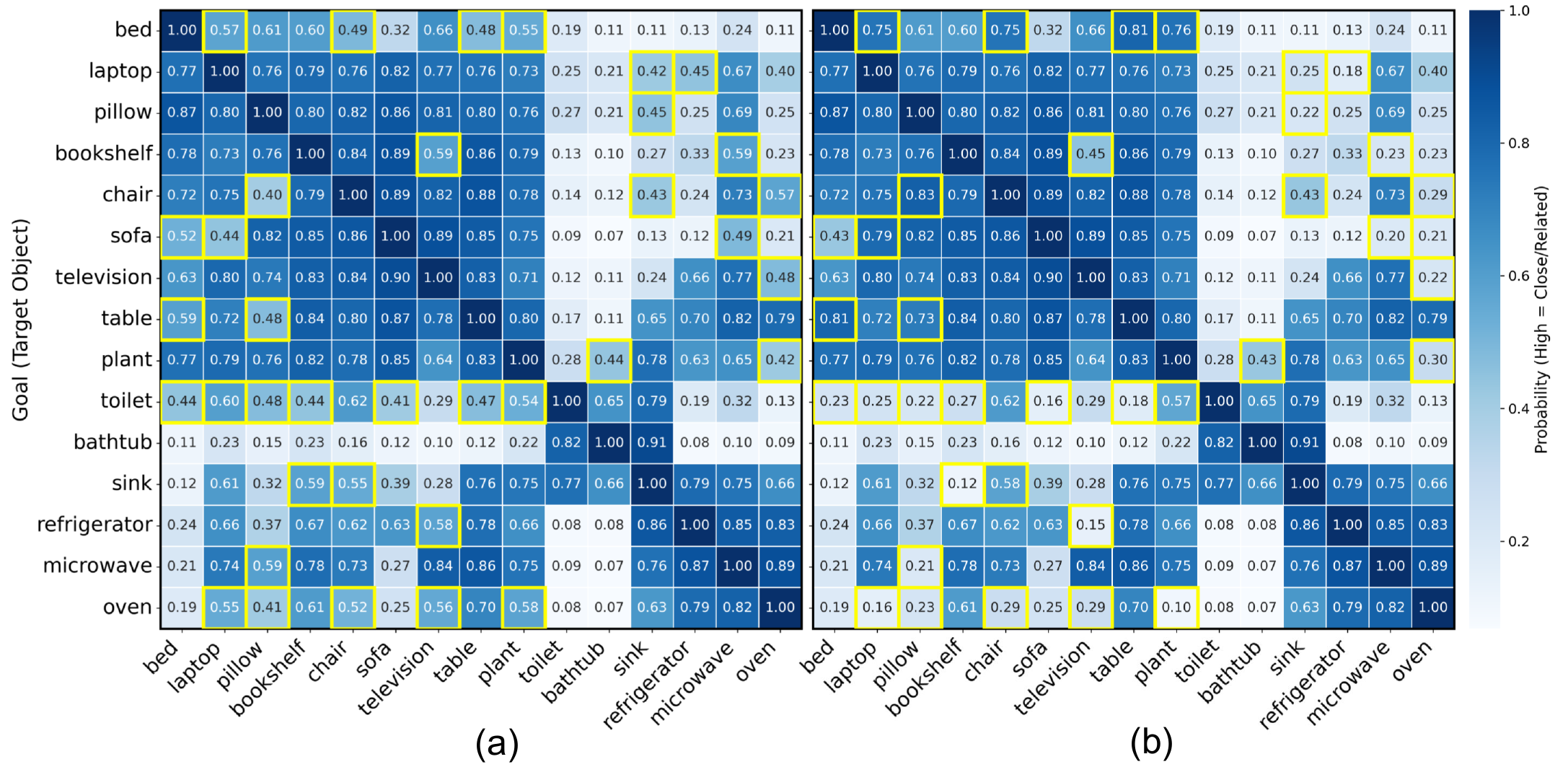}
    \caption{Comparison of Logit Distributions. (a) Uncalibrated logits from the Fast Mode showing high uncertainty. (b) Calibrated logits after adaptive slow-mode deliberation, showing a clean separation and higher confidence in the correct association.}
    \label{fig:object_calibrated_pair}
\end{figure}

We first illustrate the effectiveness of our adaptive inference mechanism in resolving ambiguity. In scenarios involving implicit spatial relationships, the Fast Mode (System 1) often generates high logits for multiple potential candidates, creating a noisy graph of associations. As observed in Figure~\ref{fig:object_calibrated_pair} (Left), the model assigns high probabilities to irrelevant pairs, such as linking ``Toilet'' with ``Laptop'' ($0.60$), implying a misleading certainty. This ambiguity causes the Conformal Prediction module to construct a large prediction set (size $>1$), thereby triggering the Slow Mode (System 2). By engaging in chain-of-thought reasoning, System 2 evaluates semantic plausibility and calibrates these logits. As shown in Figure~\ref{fig:object_calibrated_pair} (Right), this process effectively suppresses noise by reducing the (``Toilet'', ``Laptop'') score to $0.25$, while sharpening the distribution around valid cues (e.g., reinforcing the ``Bed''-``Chair'' association from $0.49$ to $0.75$). This adaptive calibration ensures that the agent follows semantically valid landmarks rather than being misled by the uncalibrated confidence of the Fast Mode.

\subsection{Failure Cases Analysis}
\label{subsec:failure_analysis}
We classify failure cases on Gibson and HM3D into three modes (Table~\ref{tab:failure_analysis}). Anomalous failed episodes are excluded from this analysis.
\textbf{Perception Failures (Type I)} dominate ($45\%$--$52\%$), stemming from domain gaps (e.g., blur, occlusion). These manifest as \textit{false negatives} (missing objects causing timeouts) or \textit{false positives} (misidentifying similar objects causing premature stops).
\textbf{Navigation Traps (Type II)} ($28\%$--$35\%$) arise from geometric complexity. Agents may get stuck in \textit{local exploration minima}, pursuing inaccessible semantic frontiers while revisit penalties hinder backtracking.
\textbf{Inference Limitations (Type III)} ($19\%$--$21\%$) reflect the risk-efficiency trade-off, including \textit{Calibration Failures} (Fast Mode is confidently wrong) and \textit{Hallucinations} (Slow Mode infers incorrect associations).

\begin{table}[htbp]
\centering
\caption{\textbf{Distribution of Failure Modes.} Values show percentage of failed episodes.}
\label{tab:failure_analysis}
\resizebox{0.48\textwidth}{!}{%
\begin{tabular}{lccc}
\toprule
Dataset & Perception & Navigation & Inference \\
\midrule
Gibson & 45.5\% & 35.2\% & 19.3\% \\
HM3D v1 & 52.3\% & 28.5\% & 19.2\% \\
HM3D v2 & 48.8\% & 30.0\% & 21.2\% \\
\bottomrule
\end{tabular}%
}

\end{table}

\section{Realworld Experiments}
\label{sec:realworld}

\begin{figure*}[t]
    \centering
    \includegraphics[width=0.8\linewidth]{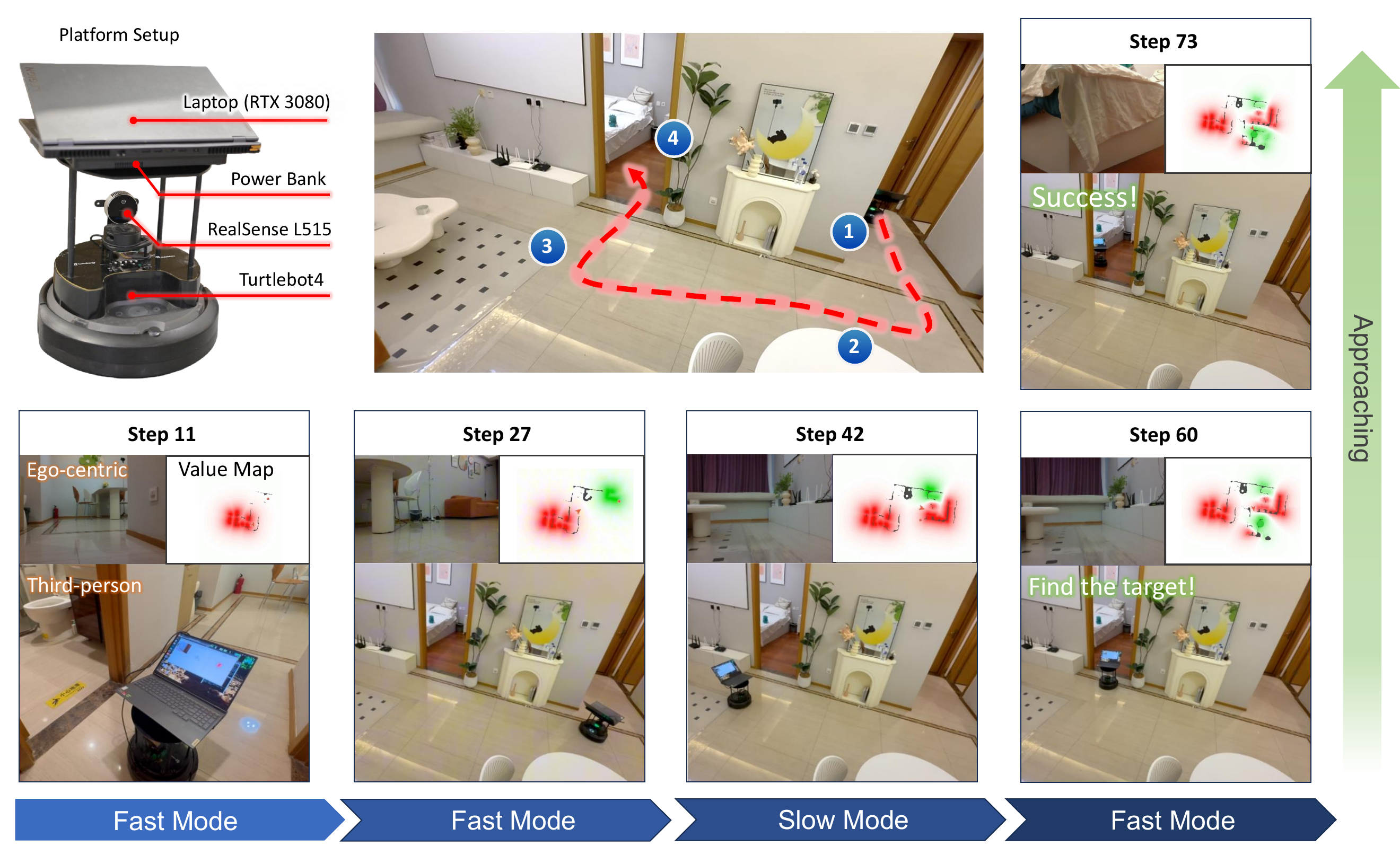}
    \caption{Visualization of Real World Experiments. The figure depicts the navigation trajectory of the TurtleBot 4 platform (setup shown in top-left) searching for a toy. The system dynamically switches between Fast and Slow modes based on uncertainty. During the initial approach (Steps 11-27), the agent operates in \textbf{Fast Mode} for efficient exploration. At Step 42, high uncertainty at a semantic transition triggers the \textbf{Slow Mode}, enabling the agent to infer the correct path into the bedroom. The agent then reverts to Fast Mode (Step 60) and successfully localizes the target (Step 73). This demonstrates the efficacy of the adaptive fast-slow inference mechanism in handling real-world ambiguities.}
    \label{fig:real-world}
\end{figure*}

To validate the efficacy of CovNav in physical environments, we conduct experiments on a real-world mobile robot platform.

{\bf Hardware Setup.} We utilize a TurtleBot 4 mobile base equipped with an Intel RealSense L515 LiDAR camera, which replaces the default sensor to provide high-fidelity RGB and depth observations. To enable fully autonomous onboard computing, we mount a laptop with an NVIDIA RTX 3080 GPU, powered by an external battery, to run all perception and LLM inference processes locally.

{\bf Perception.} The L515 sensor captures aligned RGB and depth images at a resolution of $640 \times 480$. We align the Horizontal Field of View (HFOV) to $56^{\circ}$ to match the simulation settings. The camera is mounted at a height of $0.18$m relative to the base, maintaining a horizontal viewing angle.

{\bf Action Space Control.} We bridge the gap between continuous real-world dynamics and the discrete action space defined in simulation (Forward, Turn Left, Turn Right, Stop) through a velocity-pulse control scheme. Using the robot's underlying PID controller, we execute ``Forward'' actions via a $0.25$m/s velocity command to approximate a $0.2$m displacement. Rotational actions are executed with an angular velocity of $\pm 0.5$rad/s, resulting in an approximate $30^{\circ}$ turn.

{\bf Algorithm Deployment.} To achieve standalone autonomy without reliance on cloud APIs, we deploy a quantized Qwen3-8B-FP8 model onboard. Additionally, the pre-computed Object Relationship Matrix serves as an onboard static memory for efficient O(1) queries. The fast-slow inference mechanism is implemented by dynamically toggling the model's generation parameters for open-set queries: the ``\texttt{enable\_thinking}'' flag is disabled for the Fast Mode to prioritize speed, and enabled for the Slow Mode to activate deliberative reasoning when uncertainty is high.

{\bf Real-world Navigation Process.} We validate our system across three diverse multi-room indoor environments, testing with various semantic targets including toilet, sofa, bed, and toy. Figure~\ref{fig:real-world} presents a representative navigation episode where the robot is tasked with finding a ``Toy''. The trajectory breakdown demonstrates the effectiveness of our adaptive fast-slow mechanism:
\begin{itemize}[leftmargin=*]
    \item \textbf{Phase I: Efficient Exploration (Steps 0-41).} The robot initiates the search in the hallway and dining area. In this open area with clear semantic landmarks, the CP module maintains a low uncertainty estimate and identifies low semantic relevance to the target ``toy''. Consequently, the agent operates in \textbf{Fast Mode}, utilizing the efficient local planner to rapidly cover ground and move through the corridor without hesitation.
    
    \item \textbf{Phase II: Slow Deliberation at Ambiguities (Step 42).} As the robot reaches a critical decision point at the doorway, the visual context becomes ambiguous (transition from corridor to living room). This semantic uncertainty triggers the \textbf{Slow Mode}. The quantized Qwen3-8B-FP8 model is re-invoked with thinking enabled to perform chain-of-thought reasoning. It successfully infers the high probability of the target's presence around the sofa and in the adjacent bedroom based on the semantic context, directing the robot to traverse the living room.
    
    \item \textbf{Phase III: Precise Localization (Steps 43-73).} Once inside the target room, the uncertainty subsides. The system reverts to \textbf{Fast Mode} to perform a fine-grained object search, successfully localizing the target at Step 73. 
\end{itemize}
The successful completion of this episode in 73 steps validates the stability of our onboard deployment. Crucially, the system only activated the computationally intensive Slow Mode during the critical transition phase (Step 42), while relying on the efficient Fast Mode for the majority of the trajectory. This selective activation strategy allows the robot to maintain real-time responsiveness on the limited perception-compute budget of the mobile platform, verifying the practical viability of CovNav in physical deployments.

\section{Conclusion}
\label{sec:conclusion}

In this work, we presented \textbf{CovNav}, a novel adaptive inference framework that bridges the gap between efficiency and capability in embodied AI. Drawing inspiration from dual-process theories, we introduced a mechanism that dynamically switches a local LLM between fast direct inference and slow deliberative reasoning modes. By leveraging Conformal Prediction to rigorously quantify uncertainty without additional training, CovNav serves as a reliable gatekeeper for this switching logic. Extensive experiments on Gibson and HM3D demonstrate that our method achieves state-of-the-art ZSON performance, matching the navigation performance of always using the slow deliberative mode while reducing latency to nearly one-third. Ultimately, CovNav offers a principled path toward trustworthy embodied agents that are self-aware of their limitations, providing valuable guidance for future research in integrating uncertainty quantification with active perception and robust neuro-symbolic planning.


\ifCLASSOPTIONcaptionsoff
  \newpage
\fi



%



\bibliographystyle{IEEEtran}
\bibliography{references}

%





\end{document}